# Fine-tuned Vision Language Model for Localization of Parasitic Eggs in Microscopic Images

Chan Hao Sien
Faculty of AI & Engineering
Multimedia University
Cyberjaya,Malaysia
1211102896@student.mmu.edu.my

Hezerul Abdul Karim*
Faculty of AI & Engineering
Multimedia University
Cyberjaya,Malaysia
hezerul@mmu.edu.my
*Corresponding author

Nouar AlDahoul*
Computer Science Department
New York University
Abu Dhabi, Unite Arab Emirates
nouar.aldahoul@nyu.edu
*Corresponding author

*Abstract*—Soil-transmitted helminth (STH) infections continuously affect a large proportion of the global population, particularly in tropical and sub-tropical regions, where access to specialized diagnostic expertise is limited. Although manual microscopic diagnosis of parasitic eggs remains the diagnostic gold standard, the approach can be labour-intensive, time-consuming, and prone to human error. This paper aims to utilize a vision language model (VLM) such as Microsoft Florence that was fine-tuned to localize all parasitic eggs within microscopic images. The preliminary results show that our localization VLM performs comparatively better than the other object detection methods, such as EfficientDet, with an mIOU of 0.94. This finding demonstrates the potential of the proposed VLM to serve as a core component of an automated framework, offering a scalable engineering solution for intelligent parasitological diagnosis.

*Keywords—Visual Language Model, Parasitic Egg, Microscopic Image, Object Detection, Classification, Reporting*

## I. INTRODUCTION

Parasitic infections remain a significant global public concern to mankind, specifically to those in tropical and subtropical regions, where there is limited access to proper sanitation and healthcare. According to the World Health Organization (WHO), over 1.5 billion people are infected with soil-transmitted helminths globally, resulting in mass disability and mortality [1].

Intestinal parasitic infections can cause a range of health issues like abdominal pain, digestive disorders, malnutrition, impaired growth, or in severe cases, intestinal obstruction. There are over 300 species of human intestinal parasites [2], and some, like Ascaris lumbricoides, reproduce at a shocking rate of 200,000 eggs per day [3]. Timely and accurate diagnosis is therefore crucial to prevent complications and to break the transmission cycle. Any delay in diagnosis can lead to rapid disease progression and increase the risk of community spread.

However, the traditional way of diagnosing parasitic infections relies heavily on manual microscopic examinations of fecal samples, and there are a few drawbacks to this approach. Firstly, the morphological similarity and distractions caused by impurities in the sample make it challenging for manual inspection of different types of parasite eggs using microscopes. To ensure accurate diagnosis, extensive training is required to ensure they gain sufficient expertise in diagnosis. On average, an expert technician takes 30 minutes to complete an examination of a single sample [4], making the process both labor-intensive and time-consuming. Secondly, the inadequate number of skillful personnel in the field often lead to fatigue and increases the likelihood of human error, making the traditional method less reliable.

Hence, developing an automated system for the identification of parasitic eggs is essential to minimize human errors associated with manual inspection of smear samples and to provide faster, more efficient, and more precise diagnosis outcomes [10, 11]. Along the way, researchers have developed computer vision systems and image recognition techniques for automatic identification of parasitic eggs in microscopic images. These solutions evolved from classical machine learning [5] to modern deep learning techniques [6, 7, 8, 9, 10, 11].

The key contributions of this paper include the following:

1. We evaluated the localization performance of base Florence-2 for localization task of parasitic eggs in microscopic images.

2. We fine-tuned Florence-2 with ICIP22 dataset that contains coordinates of parasitic eggs in microscopic images.

3. We evaluated the localization performance of fine-tuned Florence-2 and compared it with other methods.

## II. RELATED WORK

A classical machine learning framework was proposed by Ray et al. [5] for detecting Ascaris lumbricoides eggs, utilizing handcrafted features and traditional segmentation techniques. The proposed model employed ANN and SVM classifiers for classifications. Additionally, several works have explored advanced deep learning architectures for parasitic egg detection, particularly in the ICIP 2022 challenge. Bandara et al. [6] proposed a parallel framework that decouples classification and localization using dual ResNet models, while Aung et al. [7] formulated the task as a multi-task learning problem integrating object detection and instance segmentation with pseudo-mask supervision. Ensemble-based and transformer-enhanced detectors were also widely adopted [8][9], including Cascade R-CNN with Swin-Transformer backbones and DETR-based architectures, demonstrating strong performance improvements over conventional CNN-based models.

To further enhance classification accuracy, hybrid and fusion strategies have been investigated. AlDahoul et al. [10] combined EfficientDet with an auxiliary EfficientNet-based classifier, while their later work [11] introduced CoAtNet to integrate convolutional inductive bias with transformer capacity, achieving improved generalization.

## III. METHODOLGY

### A. Dataset description

A parasitic egg dataset, namely the Chula-ParasiteEgg-11 [4] dataset was proposed in the ICIP challenge 2022, which contains various types of parasitic egg species. The dataset consists of 11 categories of parasitic egg collected from faecal smear samples, where the average egg sizes in the sample generally fall within 15-100 μm. The among categories includes: A. lumbricoides, Capillaria philippinensis, Enterobius vermicularis, Fasciolopsis buski, Hookworm, Hymenolepis diminuta, H. nana, Opisthorchis viverrini, Paragonimus spp., Taenia spp., and T. trichiura. Figure 1 shows some samples of the microscopic images with different types of parasitic egg. The entire dataset contains 11000 images and 2750 images for training and testing dataset respectively, resulting in the largest collection of its kind. Image acquisition is carried out using a variety of devices, including Canon EOS 70D camera body with Olympus BX53 microscopes, iPhone 12 and 13 with either 10× times eyepieces lens of Nikon Eclipse Ni, Samsung Galaxy J7 Prime phone, DS-Fi2 Nikon camera body with Nikon Eclipse Ni microscopes or Olympus BX53 devices. This approach introduces differences in resolutions, lighting and setting conditions to the images. In addition, some images are made to be out-of-focus, exhibit noise and motion blur, having been captured with a motorized stage microscope. The broad spectrum of image quality and visual characteristic is included intentionally to improve the robustness and generation capability of the detection models. Figure 1 shows examples from this dataset.

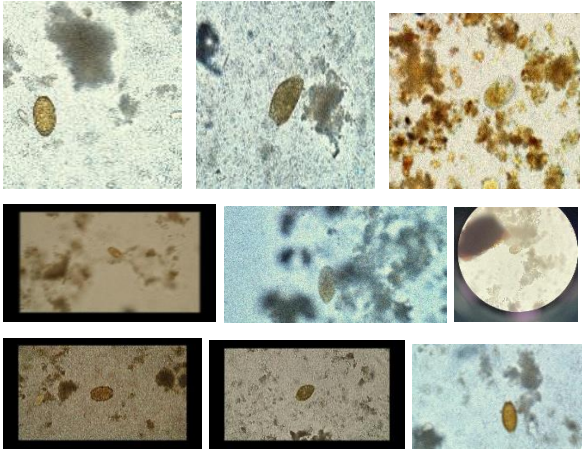

Fig.1: Illustration of microscopic image samples with variety of conditions (different resolution, clearness, etc.) in the Chula-ParasiteEgg-11 dataset

### B. Florence-2

Florence-2 [12] is a unified vision model that employs a prompt-driven representation for supporting multiple tasks related to computer vision and vision-language tasks. Unlike existing large vision models which are restricted to single vision tasks, Florence-2 was created for the capability to handle text-prompt as the model task instructions and generate designated results in test format for a variety of tasks, ranging from captioning, grounding, object detection to segmentation. Florence-2 model is built in a sequence-to-sequence design structure, consisting of two key components, the DaVit image encoder and a multimodal encoder-decoder. The model has been pretrained on the FLD-5B dataset, a dataset with 5.4 billion comprehensive visual annotations on 126 million images. In terms of performance, the Florence-2 reaches a new state-of-the-art zero-shot performance level in tasks like captioning on COCO and visual grounding on Flick30k dataset. In terms of fine-tuning capability, the model can compete with larger specialist models after the model is fine-tuned with public human-annotated data. There are two variances of Florence-2 which are available, the base model (0.23B) and the large model (0.77B). In this paper, the Florence-2 large model is adopted for fine-tuning to localize all parasitic eggs in microscopic images. Figure 2 illustrates the localization system.

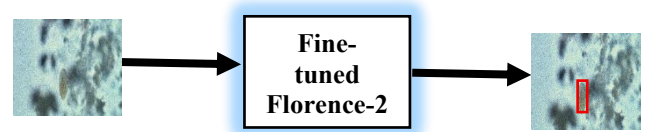

Fig.2: Illustration of the fine-tuned Florence-2 for localization of parasitic eggs.

When the microscopic image is provided as input, it first passes through a preprocessing stage where it is resized, normalized, and padded. The processed image, together with the object detection instruction prompt, is then fed into the localization vision-language model (VLM), which analyzes both inputs and outputs the predicted bounding box coordinates. These coordinates are essential for the subsequent step, as they enable cropping of the parasitic egg region from the full microscopic image for further analysis.

## IV. RESULT AND DISCUSSION

### A. Experimental Setup

The training of the proposed localization model (Florence-2) was conducted by using a Pytorch framework on a NVIDIA RTX2000 Ada GPU. The microscopic images were resized to 768x768 before being input into the Florence-2-large model. In addition, the images were being normalized. The number of epochs was set to 3. The learning rate is set to 0.00005. The batch size per device is set to 1 and the gradient accumulation steps are set to 8. For the Lora adapter, the Lora rank is set to 8.

The training set of 11,000 labeled microscopic images was used in this experiment. The entire database was divided into three subsets: 60% (6600 images) for training, 20% (2200 images) for validation and 20% for testing. The qualitative and quantitative results are shown in Figures 4 and 6 for the testing datasets.

### B. Localization result

In this section, we evaluate the localization performance of base and Fine-tuned Florence-2 and compare them with the EfficientDet detector proposed in [10]. The distributions of IoU scores between the ground-truth and predicted bounding boxes for the base and fine-tuned Florence-2 models are presented in Figures 3 and 4. The results indicate that the base Florence-2 model is unable to accurately localize parasitic eggs, likely because such images were not included in its training data. The strong spike at 0 indicates a significant number of missed or poorly localized detections, suggesting that the base model struggles with accurate object alignment without task-specific adaptation. In contrast, after fine-tuning with parasitic egg images, the model's localization performance improved substantially, achieving a high mean IoU of 0.94. A few samples of the localization capability of the fine-tuned Florence-2 are shown in Figure 6.

We compared the distributions of IoU scores between fine-tuned Florence-2 and EfficientDet proposed in [10] as shown in Figure 5. EfficientDet, while still performing well, shows a slightly broader spread across the 0.80–0.95 range and a lower peak frequency, suggesting comparatively higher variance in localization.

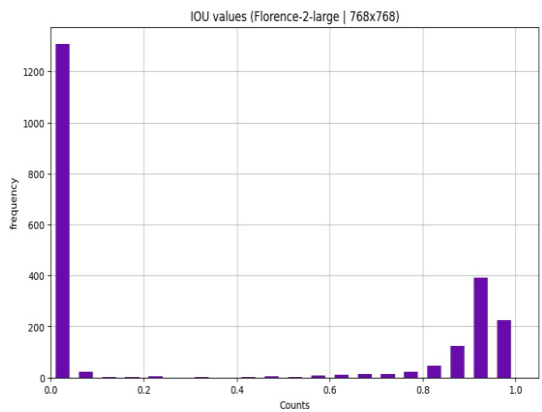

Fig.3: IOU distribution of the base Florence-2

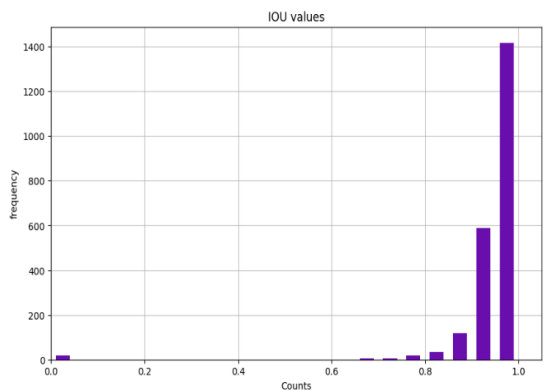

Fig.4: IOU distribution of the fine-tuned Florence-2

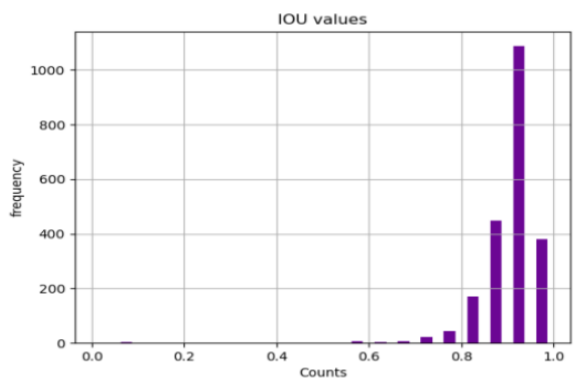

Fig.5 : IOU distribution of the EfficientDet detector proposed in [10].

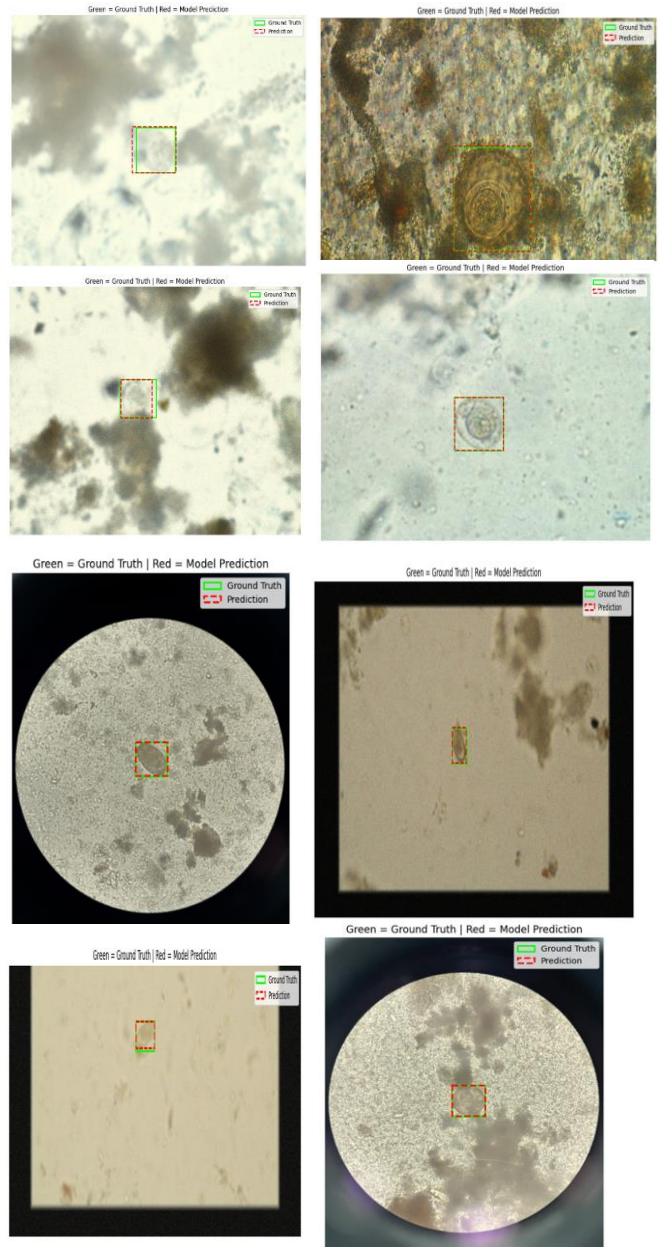

Fig.6: Illustrations of predicted bounding box by Florence-2 and the actual bounding box for detecting parasitic eggs in microscopic images with different conditions

V. CONCLUSION AND FUTURE WORK

In this paper, we utilized a vision language model for automatic localization of parasitic eggs in microscopic images. Florence-2 was fine-tuned for localization task. The qualitative results show the effectiveness of fine-tuned Florence-2 in correctly localizing parasitic eggs under varying microscopic imaging conditions. Furthermore, a comparison between the fine-tuned Floremce2 and other object detection models such as EfficientDet in terms of IOU distribution highlights superior accuracy and consistency of fine-tuned Florence-2. Its distribution shows a higher frequency of near-perfect overlaps and fewer low-IoU outliers, indicating more precise and stable bounding box regression. In contrast, Overall, the results suggest that fine-tuning Florence-2 leads to improved detection precision and more consistent object localization compared to EfficientDet and base Florence-2.